*Research Article*

# Using a novel fractional-order gradient method for CNN back-propagation

**Mundher Mohammed Taresh [1,\*], Ningbo Zhu [1], Talal Ahmed Ali Ali [1] and Weihua Guo [1]**

[1] College of Information Science and Engineering, Hunan University, Changsha, Hunan, China; quietwave@hnu.edu.cn (Zhu. N); taaw2012@hnu.edu.cn (Talal. A); wayfar@hnu.edu.cn (Guo. W)
\* Correspondence: mundhert@hnu.edu.cn (MM. Taresh); Tel.: +86-18573153530

**Abstract:** Computer-aided diagnosis tools have experienced rapid growth and development in recent years. Among all, deep learning is the most sophisticated and popular tool. In this paper, researchers propose a novel deep learning model and apply it to COVID-19 diagnosis. Our model uses the tool of fractional calculus, which has the potential to improve the performance of gradient methods. To this end, the researcher proposes a fractional-order gradient method for the back-propagation of convolutional neural networks based on the Caputo definition. However, if only the first term of the infinite series of the Caputo definition is used to approximate the fractional-order derivative, the length of the memory is truncated. Therefore, the fractional-order gradient (FGD) method with a fixed memory step and an adjustable number of terms is used to update the weights of the layers. Experiments were performed on the COVIDx dataset to demonstrate fast convergence, good accuracy, and the ability to bypass the local optimal point. We also compared the performance of the developed fractional-order neural networks and Integer- order neural networks. The results confirmed the effectiveness of our proposed model in the diagnosis of COVID-19.

## 1. Introduction

Training a convolutional neural network (CNN) is a process of adjusting its weights, which requires special algorithms. The backpropagation algorithm is the backbone of deep learning, which can be learned very quickly if the key concepts of the calculus, especially the chain rules, are understood. In short, familiarity with algorithms for adjusting weights and the evolution of optimizers during training has led to the development of reliable networks that can solve most neural network training problems.

Gradient descent (GD) is a popular approach to optimize neural network weights. System identification [1], automated control [2], image noise reduction [3], and machine learning [4] are other areas where the gradient descent approach (GD) is also used. Besides, it serves as the basis or inspiration for a considerable number of efficient algorithms [5]. The authors of [6] propose a triple gradient descent method using the subspace technique. An adaptive neural gradient descent control for nonlinear dynamical systems with chaotic behavior was presented by Liu and Lan [7]. However, the "zigzag" descent and the disadvantage of sluggish convergence near the extreme value of the traditional gradient descent method significantly affect the performance of the derived optimization algorithm, which further limits its practical applicability.

Fractional calculus has become a hot topic in artificial neural network research. Like integral calculus, fractional calculus has been around for a long time. Differentiation and integration of fractions of arbitrary order are at the heart of its functionality. Fractional calculus is used in a variety of areas, including medical imaging [8], neural networks [9,10]  image and signal processing [11,12], and many other areas of computer science and engineering. The potential of fractional calculus to improve the performance and convergence speed of ordinary gradient methods has recently been demonstrated in some papers [13–17], which is mainly owing to the significant long memory characteristic.

Unfortunately, simple replacement of the conventional derivative in GD methods with fractional-order one renders the convergence to fractional extreme point rather than real extreme point of the target. This disadvantage significantly impairs the applicability of this method.

In the literature of neural network field, Caputo's derivative is used in a fractional GD backpropagation approach to train model [18–20]. To get around the above-mention issue, the truncation and short memory principle were used to modify the fractional order gradient method [18,19]. As a result, it is shown that the fractional gradient approach is convergent to the real extreme point in the same way as the integer order method is, but with a higher convergence speed. The authors of [20] introduced the fixed memory step strategy, where the constant lower integral term is replaced by a variable integral term so that the step remains fixed. However, the authors of [20] approximated the fractional order derivative using only the first term of the infinite series of Caputo's definition. This indeed significantly truncated the length of the memory, which hindered the performance of FGD to a level just above that of the traditional methods of GD. In contrast, we approximated the fractional order derivative with an adjustable number of terms ($M$), because the higher the number of terms, the closer it is to the ideal fractional.

The goal of this work is to utilize the FGD method for training the CNN model to detect COVID-19 based on X-ray images. In this work, a novel fractional order gradient method with fixed-memory step and adjustable number of terms is proposed, which is shown to provide extra degree of freedom training CNN models. The proposed FGD method is applied to the backward propagation (BP) of CNN for the identification of COVID -19 radiographs, which is a new trend in the medical image classification society. To the best of our knowledge, this is the first work to classify COVID -19 from X-ray images by training a neural network using the fractional-order gradient method with adjustable numbers. To do that, the fractional-order gradient method is designed based on Caputo's definition of fractional-order derivatives. two types of neural network backward propagation gradients are split: gradients that are transmitted across layers (interlayer gradients) and gradients that are used to update parameters inside layers (intralayer gradients). The intralayer gradients are replaced by fractional order ones, but the integer order of interlayer gradients allows the chain rule to be maintained. The CNN with fractional order gradient technique could be accomplished by connecting all layers end-to-end and adding loss functions. Furthermore, the proposed neural networks verify that fractional order gradients execute remarkable performances in rapid convergence, high accuracy, and the ability to avoid the near-optimal point. The main contribution of this work can be summarized as follows:

- We propose a new convergence solution design based on Caputo definition.
- We adopt the proposed algorithm with fixed memory step and adjustable number of terms to address the long memory truncation.
- We investigate the backward propagation of CNN by the proposed algorithm.
- We evaluate the proposed method with a complicated CNN structure.

The remainder of this paper is organized as follows. Section 2 describes the algorithm proposed for this study and provides some basic information for later use. The recommended fractional order gradient method for backpropagation of CNN is presented in Section 3. The experiments performed are presented in Section 4. Section 5 illustrates the validity of the proposed approach. Finally, the conclusions of this work are presented in Section 6.

## 2. Study Algorithm

Fractional calculus is the operation of extending derivatives and integrals to other fractions. It provides a more precise tool for describing physical systems. By Caputo's definition, the fractional derivative of a constant function is equal to 0, which is consistent with integer-order calculus. Therefore, Caputo's definition is widely used in solving engineering problems. The Caputo differential is one of its most common definitions. For a function $f$ that is defined in $[h_0, h]$, Fractional derivative to a real order $(\alpha)$ can be expressed using Caputo's definition as

$$^{Caputo}_{h_0}\mathcal{D}_h^\alpha f(h) = \frac{1}{\Gamma(r-\alpha)} \int_{h_0}^{h} (h-\tau)^{r-\alpha-1} f^{(r)}(\tau) d\tau, \qquad (1)$$

where $\Gamma(\alpha) = \int_0^\infty k^{\alpha-1} e^{-k} dk$ is the gamma function, $\alpha \in [r-1, r]$, and $r$ is a positive integer near $\alpha$. In discrete form, (1) could be formulated as

$$^{Caputo}_{h_0}\mathcal{D}_h^\alpha f(h) = \sum_{v=r}^{\infty} \frac{f^{(v)}(h_0)}{\Gamma(v+1-\alpha)} (h-h_0)^{(v-\alpha)}. \qquad (2)$$

When sufficient finite terms are available to adequately represent $f(\cdot)$, Equation (2) can be used efficiently and effectively. Assume that $f(k)$ is a convex smooth function with a unique extreme point $k^*$. The traditional gradient method is expressed by the update rule as $k_{\eta+1} = x_\eta - \mu f^{(1)}(k_\eta)$, where µ > 0 is the learning rate and $\eta$ denotes the number of iterations. In contrast, the fractional order gradient method is formulated differently as shown in (3).

$$k_{\eta+1} = k_\eta - \mu \,^{coputo}_{k_0}\mathcal{D}_{k_\eta}^\alpha f(k). \qquad (3)$$

In case of the fractional order derivatives are directly applied in (3), the fractional order gradient method converges to an extreme point specified by the fractional order derivatives definition, which is different from the real extreme point $k^*$ [17]. In order to converge to an extreme point, with $0 < \alpha < 1$, the update rule is replaced by

$$k_{\eta+1} = k_\eta - \mu \,^{coputo}_{k_{\eta-1}}\mathcal{D}_{k_\eta}^\alpha f(k), \qquad (4)$$

where

$$^{coputo}_{k_{\eta-1}}\mathcal{D}_{k_\eta}^\alpha f(k) = \sum_{v=1}^{\infty} \frac{f^{(v)}(k_{\eta-1})}{\Gamma(v+1-\alpha)} (k_\eta - k_{\eta-1})^{(v-\alpha)}. \qquad (5)$$

It is worth noting that the arithmetic complexity in equation (5) is acceptable if only finite terms are nonzero or their approximation is sufficient. Therefore, according to (2) and by truncating the upper terms $(v > M)$, the update rule is expressed as

$$k_{\eta+1} = k_\eta - \mu \sum_{v=1}^{M} \frac{f^{(v)}(k_{\eta-1})}{\Gamma(v+1-\alpha)} (|k_\eta - k_{\eta-1}|+\varphi)^{(v-\alpha)}. \qquad (6)$$

Where φ is a small number that introduced to avoid the non-convergence when $k_\eta = k_{\eta-1}$. It has been observed that such method converges to a real extreme point as fast as traditional one. In this paper, the proposed deep neural network is trained on Caputo differential equation, in which its derivative of constant is equal to zero. Further, it is worth mentioning that the fractional-order derivative has a special long-memory characteristic, which will play a pivotal role in the fractional-order gradient method.

## 3. CNN with fractional order gradients

Forward propagation (or forward pass) refers to the calculation and storage of intermediate variables (including outputs) for a neural network from the input layer to the output layer in order. we pose a fractional-order CNN with $X$ layers, $n^{(l)}$ ($l = 1, 2, \ldots X$) is the node number of the $l^{th}$ layer, j = 1,2,3,......,$n^{[l]}$, and $w^{[l]} = (w_1^{[l]}, w_2^{[l]}, \ldots, w_{n^{(l-1)}}^{[l]}) \in \mathbb{R}^{n^{(l-1)}}$ is the weight. We consider $a^{[l]} = (a_1^{[l]}, a_2^{[l]}, \ldots, a_{n^{(l)}}^{[l]})$ as the internal output of the $l^{th}$ layer. In order to calculate the net input $y^{[l]} = (y_1^{[l]}, y_2^{[l]}, y_3^{[l]}, \ldots \ldots y_{n^{(l)}}^{[l]})$ of the $l^{th}$ layer, we add the bias $b^{[l]} \in \mathbb{R}$ to the product of each internal input $a^{[l-1]}$ by the weight $w^{[l]}$. Thus, the procedure the forward propagation for the fractional-order deep CNN is computed as

$$y^{[l]} = w^{[l]} \cdot a^{[l-1]} + b^{[l]}. \qquad (7)$$

We then squash $y^{[l]}$ using the rectified linear unit (ReLU) function $f$ to get the $a^{[l]}$.

Using a sequence of convolutional filters, convolution reveals the input image and activates the associated features in the image. Mapping negative values to zeros and preserving positive values is made possible by ReLU. It is also referred to as activation since only the activated features are carried over to the next layer. Since our goal is to classify the positive (COVID-19) and negative (non-COVID) cases of X-ray images, we use binary cross-entropy as a loss function that can be expressed as

$$L = -\frac{1}{m}\sum_{s=1}^{m} t_s \log f(\hat{t}_s), \qquad (8)$$

where the subscript $(m)$ indicates the batch size and $s^{th}$ denotes the number of samples in a batch. $t_s$ is the label with one-hot form and $f(\hat{t}_s)$ is the sigmoid function that activated the last layer corresponding to the CNN score $\hat{t}_s$ for each $s^{th}$ sample, which given by $\frac{1}{1+e^{-\hat{t}_s}}$.

In order to minimize the total error of the fractional-order CNN, the weights are updated by the fractional gradient descent method with Caputo derivative. The gradients of the backward propagation are a mixture of fractional and integer order caused by the imperfect application of the chain rule to fractional order derivatives. To allow continuous use of the chain rule, the gradient is propagated across levels in integer order (transferring gradient), while the updated gradient is used in fractional order (updating gradient). Figure 1 shows the forwarding propagation and backward propagation for a single hidden neuron $i$ in $l^{th}$ layer.

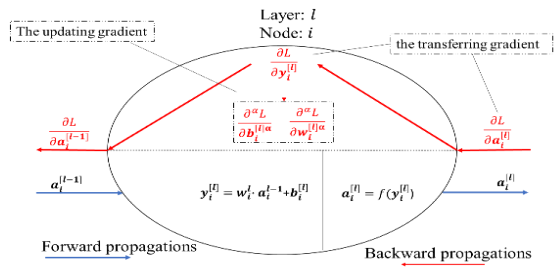

**Figure 1.** Backward and forward propagation of a single hidden neuron.

We defined $\alpha$ as the fractional-order and $\frac{\partial^\alpha L}{\partial w^{[l]\alpha}}$ and $\frac{\partial^\alpha L}{\partial b^{[l]\alpha}}$ as fractional order gradients of $w^{[l]}$ and $b^{[l]}$, respectively, which are given by

$$\frac{\partial^\alpha L}{\partial w^{[l]\alpha}} = \sum_{v=1}^{M} \frac{f^v\left(w_{(\eta-1)}^{[l]}\right)}{\Gamma(v+1-\alpha)} (|w_{(\eta)}^{[l]} - w_{(\eta-1)}^{[l]}| + \varphi)^{(v-\alpha)},$$

$$\frac{\partial^\alpha L}{\partial b^{[l]\alpha}} = \sum_{v=1}^{M} \frac{f^v(b_{(\eta-1)}^{[l]})}{\Gamma(v+1-\alpha)} (|b_{(\eta)}^{[l]} - b_{(\eta-1)}^{[l]}| + \varphi)^{(v-\alpha)}.$$

(9)

Then, the weights and biases are updated as follows:

$$w_{(\eta+1)}^{[l]} = w_{(\eta)}^{[l]} - \mu \frac{\partial^\alpha L}{\partial w^{[l]\alpha}},$$

$$b_{(\eta+1)}^{[l]} = b_{(\eta)}^{[l]} - \mu \frac{\partial^\alpha L}{\partial b^{[l]\alpha}}.$$

(10)

As modified fractional order gradient (6) is applied smoothly, the speed of convergence is improved and real extreme point can be reached now.

## 4. Experiments

In this section, a brief description of the material and parameters used in this experiment are explained. The task of this experiment is to use fractional order convolutional neural networks to identify covid-19 in a CXR image.

*4.1 Dataset*

This experiment is based on the COVID -Net Open-Source Initiative - COVIDx V9B dataset. The COVIDx V9B is an open-access benchmark dataset created by [21], and was downloaded from https://github.com/lindawangg/ COVID -Net/blob/master/docs/COVIDx.md, on 11-1-2022. It contains 30,882 CXR images from a multinational cohort of over 16,400 patients. The dataset is divided into two classes (Figure 2): COVID -19, which contains 16,690 images, and non-COVID, which includes 14192 images. 21477 images are used for training, while the rest is used for testing. The chest X-ray images are resized to 224x224 pixels before being fed into the CNN.

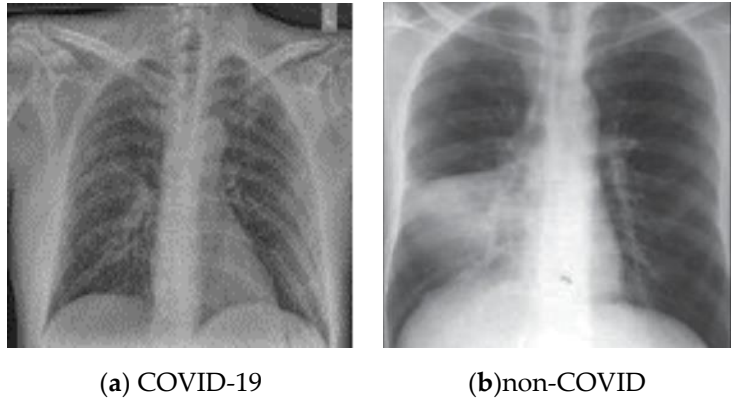

(**a**) COVID-19　　　　　　(**b**)non-COVID

**Figure 2.** Samples of X-ray images used in this study.

*3.1 Deep learning model*

In this study, the convolutional neural network model known as Visual Geometry Group-16 (VGG-16) [22] is used. Figure 3 shows the schematic of the VGG-16 used. The VGG-16 framework consists of 16 weighted layers. Of these, 13 convolutional layers are distributed in five blocks and the rest are three fully connected layers. Each block starts with convolutional layers followed by the pooling layer, which has lower position sensitivity but higher capacity for general recognition [23]. The convolutional layers have weights and biases that need to be trained, while the pooling layers transform activation using (ReLU). The main distinguishing feature of VGG-16 is that it does not rely on a huge number of hyperparameters, but uses convolutional layers of a 3x3 filter with stride 1 and always the same padding and max-pooling layer of a 2x2 filter with stride 2. It is widely considered to be one of the best architectures for image processing models to date. The convolutional blocks are connected to three fully connected layers, two of which are hidden and one of which is output. The output layer is composed of nodes that reflect the total number of classes and the sigmoid activation function directly. We update the weights using momentum stochastic gradient descent optimization (learning coefficient = 0.0005, momentum = 0.9). For this purpose, the developed prediction model and training were performed by on MATLAB 9.0, using the CPU of Core Intel (TM) i7-8700K with 8GB RAM.

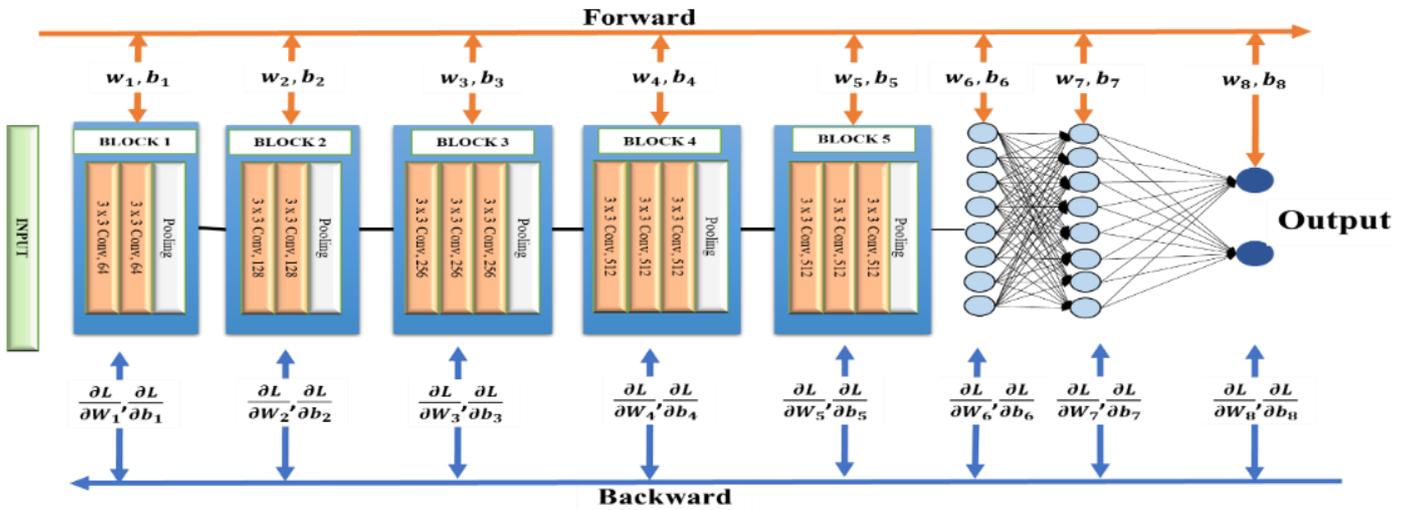

**Figure 3.** Overall architecture of the Visual Geometry Group—16 (VGG-16) model.

## 5. Results and Discussion

The experiments demonstrate the power of the proposed method and integer-order neural network. Motivated by the long memory property of fractional order derivatives, we modify the fractional order gradient method by using an adjustable number of terms (M=1,2,3,4) instead of the infinite series. Accuracy and loss are the main metrics used to measure the two different networks. Each network was trained 6 times and the average values were recorded. Some parameters were randomly initialized each time, such as the weights ∈ [-0.1,0.1], the bias ∈ [-0.1,0.1], and the input order of the samples, while the batch size m = 10, the epoch = 6, and the iteration = 114 were reserved for all experiments. Figure 4 shows the average of training and testing accuracy for each M with different fractional orders ($\alpha$= 0.1, 0.3, 0.5, 0.7, 0.9).

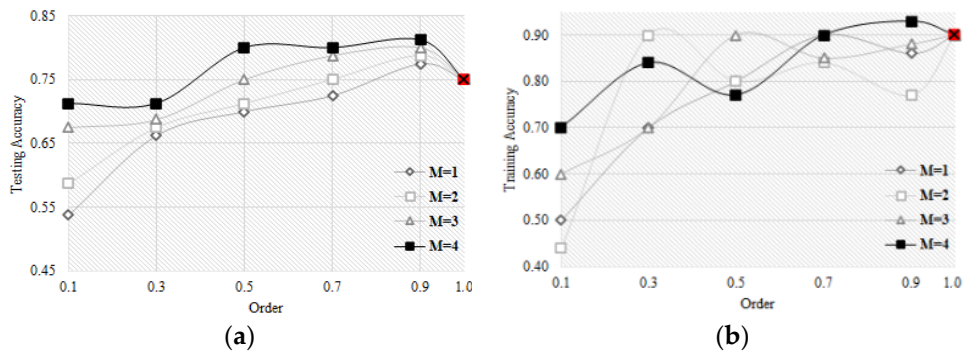

(a)        (b)

**Figure 4.** The performance comparison of different $\alpha$ for $M$: (a) The average testing accuracy; (b) The average training accuracy.

It is clear to see that the CNN works well at $\alpha \geq 0.7$ in terms of training and testing accuracy compared to integer order $\alpha =1$. Since the fractional calculus has an extremely large gamma function for $\alpha < 0.5$, all values are low. However, some unacceptable results occurred at $\alpha=0.7$, especially with M=1 and 2. As the result, the training and testing accuracy with $\alpha=0.7$ and 0.9 are shown in Table 1.

**Table 1.** The average training and testing accuracy (%).

| The number of terms | $\alpha = 0.7$ | | $\alpha = 0.9$ | |
|---|---|---|---|---|
| | Training | Testing | Training | Testing |
| M=1 | 90 | 72.5 | 86 | 77.5 |
| M=2 | 84 | 75 | 77 | 78.75 |
| M=3 | 85 | 78.75 | 88 | 80 |
| M=4 | 90 | 80 | 93 | 81.25 |

The average of training and testing accuracy for $\alpha = 1$ is **90%** and **75%**, respectively.

The results shown in Table 1 prove that gradient methods of fractional order (a > 0.7) work well with CNN. The best values of fractional order gradient methods are obtained of both with M=4, and their values exceed the value obtained by the integer-order gradient method. In general, the testing accuracy of the CNN improves with fractional order as the number of terms M increases. At this point, it is worth noting that the proposed method at M = 1 is completely identical to the method proposed by [20] and therefore the latter is superior to the method used in [20].

To further evaluate fractional-order gradients, Figure 5 shows the comparison of the training and testing average accuracy for M=4 during iteration.

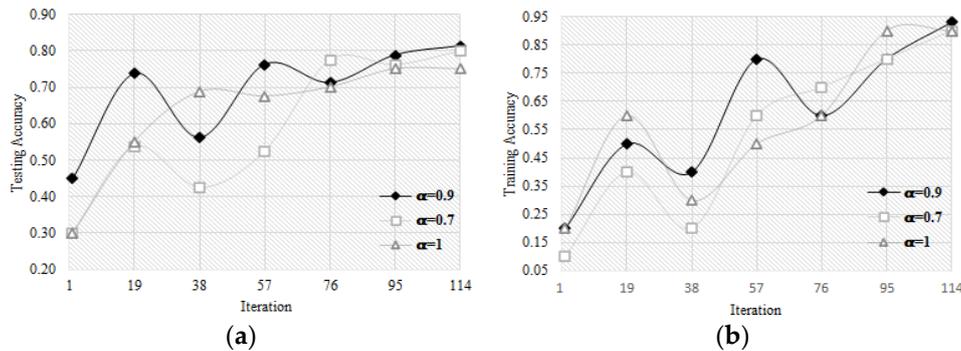

(a) (b)

**Figure 5.** The performance comparison of $\alpha$=0.7, 0.9, and 1 for M=4: (a) The average testing accuracy; (b) The average training accuracy.

Furthermore, the stability and convergence of the proposed fractional order CNN is shown in Figure 6. It shows the change of the loss values for M=4 during the iteration.

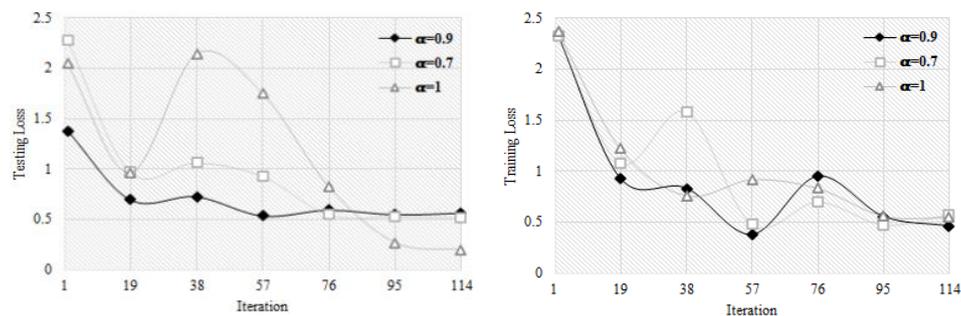

(a) (b)

**Figure 6.** The change of values of the loss for M=4 with $\alpha$=0.7, 0.9 during the iteration: (a) The average testing loss; (b) The average training loss.

Figure 6 shows that the average loss of fractional order decreases during training. However, it appears that the loss repeated by the gradients of the fractional-order favors jumping on. A larger variance usually indicates a looser distribution, which means that the fractional order gradient approach helps to optimize frequent and broad process jumps. As a result, the probability of escaping the local optimal point increases.

The training times given in Table 2 indicate the average running time for each given order. It clearly shows that the training process takes a relatively similar time for the different fractionated orders. The training time required by integer order CNN is only 0.15% less than that required by fractional-order CNN with $\alpha = 0.9$. We can also note that in other cases such as $\alpha = 0.7$ and $0.5$, there are relatively similar speeds.

**Table 2:** The training and testing loss (S).

|  | **Fractional order CNN, M=4** | | | | | **Integer order CNN** |
|---|---|---|---|---|---|---|
| Order ($\alpha$) | 0.1 | 0.3 | 0.5 | 0.7 | 0.9 | 1 |
| Time Elapsed | 44.53 | 44.36 | 37.34 | 37.57 | 42.51 | 36.35 |

The high speed of the fractional order gradient method for CNN is due to the fact that only the update gradients are replaced by the fractional order and the fractional order update gradients are obtained corresponding to the integer order gradients and the additional calculation of the fractional order updating gradients is quite simple. In this experiment, we trained the CNN model based on fractional order gradient with a fixed memory step and an adjustable number of terms to avoid memory length truncation. The promising results show that the proposed method converges to real extreme points. However, more iterations are required for the training process. Moreover, many experiments can be conducted to further validate the performance, e.g., by changing the batch size and learning rate.

## 6. Conclusions

A novel algorithm has been used to modify the performance of fractional order gradients hindered by long memory characteristics. A novel fractional-order gradient method with a fixed memory step and an adjustable number of terms is proposed. The proposed algorithm is applied to the backward propagation of the CNN using the Caputo definition of fractional-order derivatives. For this purpose, all layers have been connected end-to-end to achieve the proposed algorithm with CNN. Except for the output layer, which uses the sigmoid function, all layers are activated by the ReLU function. The results prove that there is no computational complexity when the fractional order gradient method is combined with more complex neural network structures. The VGG16 neural network trained with the COVIDx dataset was used as the basis. The results showed that as the number of terms increased, the use of fractional order gradients with CNN outperformed the use of integer-order gradients with CNN. Moreover, the proposed fractional-order gradient method confirms its fast convergence, high accuracy, and ability to bypass local optimal in neural networks compared to the integer-order case. The presented work opens new horizons for the study of the gradient method and its application with neural networks. Moreover, the scope of the available fractional-order domain is still in mind and should be investigated in future works. The generalization of the proposed neural network on unseen data is the linchpin of future work.

**Supplementary Materials:** The following supporting information can be downloaded at: www.mdpi.com/xxx/s1, Figure S1: title; Table S1: title; Video S1: title.

**Author Contributions:** Author Contributions: MM.T.: conceptualization, methodology, software, writing—original draft preparation, writing—reviewing and editing; TMA.A.: visualization, investigation; G.W.: software, validation, data curation; N.Z.: conceptualization, supervision.

All authors have read and agreed to the published version of the manuscript.

**Funding:** This research received no external funding.

**Data Availability Statement:** The data underlying this article are available in the article.

**Acknowledgments:** This work was supported by the National Natural Science Foundation [61572177]. No additional external funding received for this study.

**Conflicts of Interest:** The authors certify that there is no conflict of interest in the subject matter discussed in this manuscript.